\pdfoutput=1

\documentclass[11pt]{article}

\usepackage[preprint]{coling}

\usepackage{times}
\usepackage{latexsym}
\usepackage{url}
\usepackage[T1]{fontenc}
\usepackage[utf8]{inputenc}
\usepackage{microtype}
\usepackage{inconsolata}
\usepackage{graphicx}
\usepackage{multirow}
\usepackage{booktabs}

\title{ViSoLex: An Open-Source Repository for Vietnamese Social Media \\ Lexical Normalization}

\author{\textbf{Anh Thi-Hoang Nguyen\textsuperscript{1,2}},
  \textbf{Dung Ha Nguyen\textsuperscript{1,2}},
  \textbf{Kiet Van Nguyen\textsuperscript{1,2,*}}
\\  
  \textsuperscript{1}University of Information Technology, Ho Chi Minh City, Vietnam
\\
  \textsuperscript{2}Vietnam National University, Ho Chi Minh City, Vietnam
\\
  \small{
    \textbf{*Correspondence:} \href{mailto:kietnv@uit.edu.vn}{kietnv@uit.edu.vn}}
\\
    \small{
    \textbf{Contributing authors:} \href{mailto:anhnth@uit.edu.vn}{anhnth@uit.edu.vn},  \href{mailto:dungngh@uit.edu.vn} {dungngh@uit.edu.vn}
}
}

\begin{document}
\maketitle
\pagestyle{empty}
\begin{abstract}
ViSoLex is an open-source system designed to address the unique challenges of lexical normalization for Vietnamese social media text. The platform provides two core services: Non-Standard Word (NSW) Lookup and Lexical Normalization, enabling users to retrieve standard forms of informal language and standardize text containing NSWs. ViSoLex's architecture integrates pre-trained language models and weakly supervised learning techniques to ensure accurate and efficient normalization, overcoming the scarcity of labeled data in Vietnamese. This paper details the system's design, functionality, and its applications for researchers and non-technical users. Additionally, ViSoLex offers a flexible, customizable framework that can be adapted to various datasets and research requirements. By publishing the source code, ViSoLex aims to contribute to the development of more robust Vietnamese natural language processing tools and encourage further research in lexical normalization. Future directions include expanding the system's capabilities for additional languages and improving the handling of more complex non-standard linguistic patterns.
\end{abstract}

\section{Introduction}

The increasing presence of Non-Standard Words (NSWs) in social media has introduced significant challenges for natural language processing (NLP) systems. In Vietnamese, these challenges are particularly pronounced due to the informal, abbreviated, and non-canonical nature of social media language. Lexical normalization, which transforms NSWs into their standard forms, is essential for improving the performance of downstream tasks such as sentiment analysis, hate speech detection, and machine translation. While research on lexical normalization has made significant advancements globally, Vietnamese has lagged behind due to a lack of resources and standardized datasets.

To address these challenges, we introduce ViSoLex\footnote{\url{https://github.com/HaDung2002/visolex}}, an open-source repository for Vietnamese lexical normalization. ViSoLex provides a comprehensive solution by integrating multitask learning capabilities to simultaneously detect and normalize NSWs. This is achieved by leveraging pre-trained language models and weak supervision techniques, reducing the dependency on extensive manual labeling. Furthermore, ViSoLex incorporates a growing dictionary of NSWs for dictionary lookup, enabling efficient identification and normalization of non-standard words.

The repository is designed to address the unique linguistic challenges of Vietnamese social media text and fosters customization, allowing researchers to adapt the system for various datasets and languages. By offering a scalable and open-source solution, ViSoLex supports broader research and practical applications, advancing the field of Vietnamese NLP. This paper presents the system architecture, multitask training framework, and the extensive efforts made to improve the quality of Vietnamese NLP tasks through lexical normalization.

\section{Related Works}

The study of lexical normalization has seen significant advancements worldwide, especially in addressing the challenges of non-standard text. Early approaches like the Abbreviation Expander by \citet{ciosici-assent-2018-abbreviation} tackled abbreviation expansion in technical documents, providing a web-based solution for easy understanding of domain-specific terms. In 2019, MoNoise by \citet{van-der-goot-monoise-2019} was introduced for vocabulary normalization, using spelling correction and word embeddings with a Feature-Based Random Forest Classifier. Initially for English, it later expanded to support multiple languages, becoming a widely used multilingual tool. Furthermore, \citet{muller-etal-2019-enhancing} marked a shift toward using pre-trained language models for handling noisy text in user-generated content, framing normalization as a token prediction task. \citet{nguyen-etal-2021-learning} introduced the idea of capturing not just lexical meaning but also social context, a key aspect in understanding informal and non-canonical language. These developments paved the way for more robust normalization systems across various languages, demonstrating the potential of combining linguistic insights with modern NLP techniques. 

In the Vietnamese context, \citet{9530818} employed deep learning models like Bidirectional-GRU to solve the problem of missing diacritics. \citet{10.1007/978-3-030-89363-7_20} further advanced the field by using VSEC model, a Transformer-based approach, to correct Vietnamese spelling errors, significantly improving upon prior methods. \citet{nguyen-etal-2024-vilexnorm} introduced the first corpus, called ViLexNorm, for Vietnamese lexical normalization, a critical resource for normalizing social media text and improving downstream tasks. Additionally, \citet{10.1007/978-3-031-64779-6_1} introduced a Seq2Seq approach for normalizing NSWs, with a publicly available dataset for further research. Building upon this foundation, \citet{nguyen2024weaklysuperviseddatalabeling} proposed a novel framework that integrates semi-supervised learning with weak supervision techniques, leveraging pre-trained language models to enhance dataset quality, reduce manual labeling efforts, and normalize NSWs.

In this paper, we further advance our previous work proposed in \citet{nguyen2024weaklysuperviseddatalabeling}, now named ViSoLex, by incorporating multitask learning capabilities to simultaneously detect and normalize NSWs. Additionally, we have integrated a dictionary lookup feature for non-standard word detection. The ViSoLex repository is designed as an open-source solution for Vietnamese lexical normalization, specifically addressing the unique linguistic challenges presented by social media text, and is made publicly available for broader use and development.

\section{ViSoLex: Vietnamese Social Media Lexical Normalization}

\subsection{System Architecture}

\begin{figure}[ht]
    \centering
    \includegraphics[width=\linewidth]{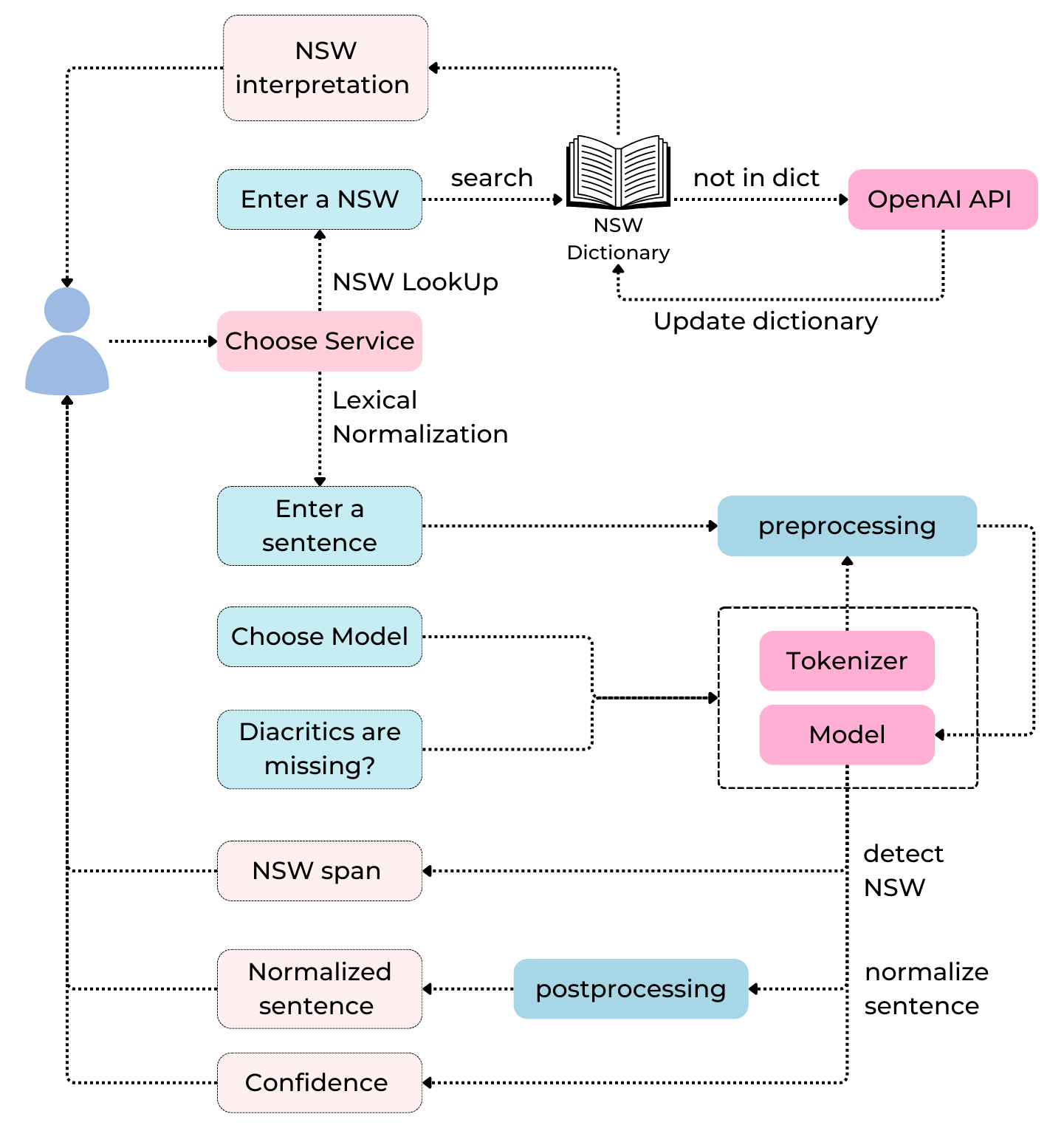}
    \caption{The Architecture of ViSoLex. The diagram illustrates the modular components enabling NSW Lookup and Lexical Normalization services, including their interactions and flow of user inputs.}
    \label{fig:architecture}
\end{figure}

ViSoLex is designed to provide two key services: NSW Lookup and Lexical Normalization. Users can input a NSW for interpretation or enter a sentence containing NSWs for normalization. The architecture of ViSoLex, as illustrated in Figure \ref{fig:architecture}, follows a modular design that integrates various components to streamline Vietnamese social media text normalization. At the core, user inputs flow through distinct paths depending on the requested service. Communication between the components ensures dynamic interaction and updates. This architecture enables independent updates to different system components while maintaining overall functionality.

\subsubsection{NSW LookUp Service}

The NSW LookUp service enables users to retrieve potential standard forms and interpretations of NSWs from an established dictionary. Upon choosing this service, users are asked to input an NSW, which is then checked against the existing dictionary. If found, the system returns the standard forms and definitions, along with relevant examples. If the word is not in the dictionary, the system consults the OpenAI GPT-4o API to suggest a possible normalization, which is then added to the dictionary for future use. This approach allows NSWs to be resolved either by utilizing existing data or dynamically learning from external models.

The NSW dictionary was built by leveraging the OpenAI GPT-4o API to generate definitions and examples for each entry in the Vietnamese Non-Standard Words Dictionary\footnote{\url{https://github.com/AnhHoang0529/vn-nsw-dictionary}}.

\subsubsection{Lexical Normalization Service}

The lexical normalization service transforms NSWs in a sentence into their standard forms. When users select this service, they input a sentence that may contain NSWs. The system tokenizes and preprocesses the input before applying a multitask-trained model (discussed in detail in Section \ref{sec:model_train}) to identify non-standard tokens and predict their corresponding standard forms, each accompanied by confidence scores. The predicted output undergoes post-processing, where redundant spaces before punctuation are removed, and proper sentence capitalization and punctuation are applied. The final result provides a fully normalized sentence, along with a breakdown of each NSW, its standard equivalent, and the confidence score, ensuring precise normalization for Vietnamese social media text.

\subsection{Lexical Normalizer Training} \label{sec:model_train}

\begin{figure}[ht]
    \centering
    \includegraphics[width=\linewidth]{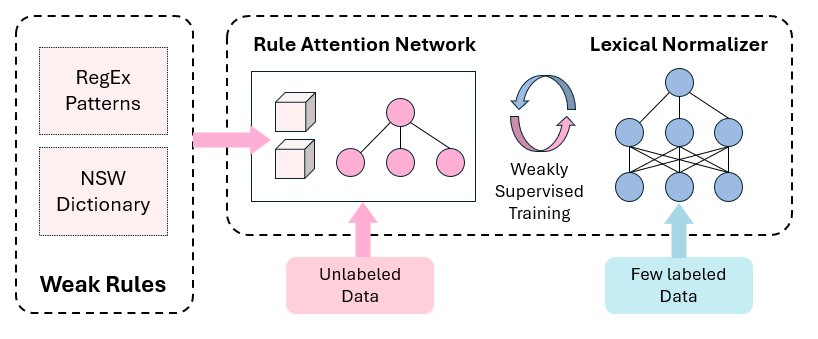}
    \caption{Weak Supervision Training. This figure illustrates the training process of the lexical normalizer, which integrates multitask learning and a Rule Attention Network guided by weak supervision rules to effectively standardize NSWs in Vietnamese social media text.}
    \label{fig:pws}
\end{figure}

The updated lexical normalizer builds on the framework presented in our previous work \citet{nguyen2024weaklysuperviseddatalabeling}, introducing multitask learning to enhance its capabilities. As illustrated in Figure \ref{fig:pws}, this weakly supervised framework leverages both labeled and unlabeled data to identify and standardize NSWs in Vietnamese social media text. Inspired by the ASTRA framework \citet{karamanolakis-etal-2021-self}, it incorporates two key components: the Lexical Normalizer, now enhanced with multitask learning as a student model, and a Rule Attention Network, acting as a teacher by embedding weak supervision rules. This integration of data-driven and rule-based approaches enables the model to generalize more effectively, handling the diverse and evolving NSW patterns in social media discourse.

\subsubsection{Lexical Normalizer}

The Lexical Normalizer is trained using a multitask framework to predict the standard forms of NSWs in Vietnamese social media text. It leverages pre-trained models, such as BARTpho \citet{tran2022bartphopretrainedsequencetosequencemodels} and ViSoBERT \citet{nguyen-etal-2023-visobert}, fine-tuned for text normalization. The input consists of sentences with NSWs, and the model outputs both NSW detection and their normalized forms, combining token classification with sequence generation for effective normalization.

ViSoLex introduces multitask learning to simultaneously handle NSW detection and lexical normalization. A shared encoder extracts the input features, followed by task-specific heads that generate predictions. The model minimizes the binary cross-entropy loss $\mathcal{L}_{NSW}$ for NSW detection and the cross-entropy loss $\mathcal{L}_{Norm}$ for normalization, with the total loss.

\begin{equation}
    \mathcal{L}_{Total} = \alpha \mathcal{L}_{Norm} + \beta \mathcal{L}_{NSW}
\end{equation}

\noindent where $\alpha$ and $\beta$ balance the contributions of each task. This multitask approach enhances efficiency and performance in normalizing noisy social media text.

\subsubsection{Rule Attention Network with Weak Rules}

To further enhance the model’s performance, a Rule Attention Network (RAN) is integrated, guided by weak supervision rules. These weak rules are derived from NSW dictionary and regular expression rules, capturing common patterns of NSWs in Vietnamese social media text. As shown in Figure \ref{fig:pws}, the RAN learns to assign different levels of attention to these rules during the training process. This network dynamically weighs the influence of each rule based on the context and reliability of the prediction, allowing the model to flexibly adapt to both well-defined and ambiguous cases of NSWs. The combination of weak supervision with the rule attention mechanism allows the model to effectively learn from limited labeled data, improving both NSW detection and normalization accuracy.

\section{Evaluation}

In this section, we evaluate the performance of multitask learning in comparison to the lexical normalization approach presented by \citet{nguyen2024weaklysuperviseddatalabeling}. The evaluation employs three key metrics: the F1-score, which specifically measures the accuracy of normalizing NSWs; the Integrity Score, which assesses the preservation of words that do not require normalization; and Accuracy, which evaluates the overall correctness of the predicted sentences. These metrics are defined and explained in detail by \citet{nguyen2024weaklysuperviseddatalabeling}.

\begin{table*}[]
\centering
\begin{tabular}{cccccc}
\toprule
\multirow{2}{*}{\textbf{Metric}}          & \multirow{2}{*}{\textbf{Task}} & \multicolumn{2}{c}{\textbf{BARTpho}} & \multicolumn{2}{c}{\textbf{ViSoBERT}} \\
                                          &                                & $p = 0.0$         & $p = 1.0$        & $p = 0.0$         & $p = 1.0$   \\
\midrule
\multirow{3}{*}{\textbf{F1-score} (\%)}        & Single task                    & 84.94             & 85.64            & 75.79             & 72.19             \\
                                          & Multitask                      & 85.28             & 85.89            & 77.22             & 75.93             \\
                                          & Improvement                    & $\uparrow$0.34    & $\uparrow$0.25   & $\uparrow$1.43              & $\uparrow$3.74              \\
\midrule
\multirow{3}{*}{\textbf{Integrity Score} (\%)} & Single task                    & 98.88             & 98.62            & 98.26             & 96.92             \\
                                          & Multitask                      & 98.83             & 98.50            & 98.64             & 98.27             \\
                                          & Improvement                    & $\downarrow$0.05             & $\downarrow$0.12            & $\uparrow$0.38              & $\uparrow$1.35              \\
\midrule
\multirow{3}{*}{\textbf{Accuracy} (\%)}        & Single task                    & 96.06             & 96.16            & 95.42             & 94.42             \\
                                          & Multitask                      & 96.14             & 96.26            & 95.08             & 94.76             \\
                                          & Improvement                    & $\uparrow$0.08              & $\uparrow$0.10             & $\uparrow$0.34             & $\uparrow$0.34              \\
\bottomrule
\end{tabular}
\caption{Comparison of Single Task and Multitask Learning Performance on BARTpho and ViSoBERT models with different diacritics removal ratios ($p = 0\%$ and $p = 100\%$).}
\label{tab:res}
\end{table*}

Table \ref{tab:res} demonstrate the impact of multitask learning on model performance across different metrics, with $p$ representing the diacritics removal ratio. The results show that multitask learning consistently enhances model performance. For BARTpho, improvements in F1-score are modest, with increases of 0.34\% and 0.25\% for different diacritics removal ratios ($p = 0$ and $p = 1$). ViSoBERT, however, benefits from multitask learning, particularly when all diacritics are removed ($p = 1$), with a notable 3.74\% increase in F1-score.

In terms of Accuracy, both models see slight improvements, but again, ViSoBERT shows stronger gains, reinforcing its ability to normalize sentence with diacritics removal in traning and development dataset. Despite a small decrease in the Integrity Score for BARTpho, ViSoBERT improves, particularly in high diacritics removal scenarios.

Overall, multitask learning proves especially effective for ViSoBERT, leading to performance improvements in handling noisy text data with diacritic variations.

\section{System Demonstration}

In this section, we outline the system demonstration of ViSoLex, tailored to meet the needs of various user groups within the NLP community. We offer two distinct entry points to accommodate both technical and non-technical users.

\subsection{For Researchers and Developers}

We have published the source code on GitHub to allow researchers and developers to leverage the system’s capabilities through the following features:

\begin{itemize}
    \item \textbf{Model Training and Evaluation}: Users can utilize the top-level script \texttt{main.py} to retrain models, reproduce results, and evaluate performance. This offers comprehensive insight into the system's underlying methodologies and processes.
    \item \textbf{Demo Functionality}: For a quick overview, users can run an interactive terminal session using \texttt{demo.py}, which demonstrates the core functionalities of the system with minimal setup.
\end{itemize}

The framework is also designed for flexibility, allowing users to customize the model and its components for specific datasets and model selection, enhancing its adaptability to various research applications. Key customization options include:

\begin{itemize}
    \item \texttt{data/}: Users can replace the default data with their own, ensuring they provide three labeled data files (\texttt{train.csv}, \texttt{dev.csv}, \texttt{test.csv}) and an unlabeled data file (\texttt{unlabeled.csv}).
    \item \texttt{dict/}: Users can integrate a custom NSW dictionary to further align the framework with their specific language or domain requirements.
    \item \texttt{aligned\_tokenizer.py}: The token-level alignment tokenizer can be modified to suit the characteristics of different datasets and languages.
    \item \texttt{normalizer/model\_construction/}: New models for lexical normalization, tailored to different languages or datasets, can be added here.
    \item \texttt{project\_variables.py}: Global constants such as data directories or language-specific tokens can be modified to fit custom requirements.
    \item \texttt{arguments.py}: Users can configure additional settings for their projects and reset the default argument values.
\end{itemize}

This modular and customizable design allows researchers to tailor the system to meet their unique needs in lexical normalization tasks.

\subsection{For Non-Experts}

\begin{figure}[ht]
    \centering
    \includegraphics[width=\linewidth]{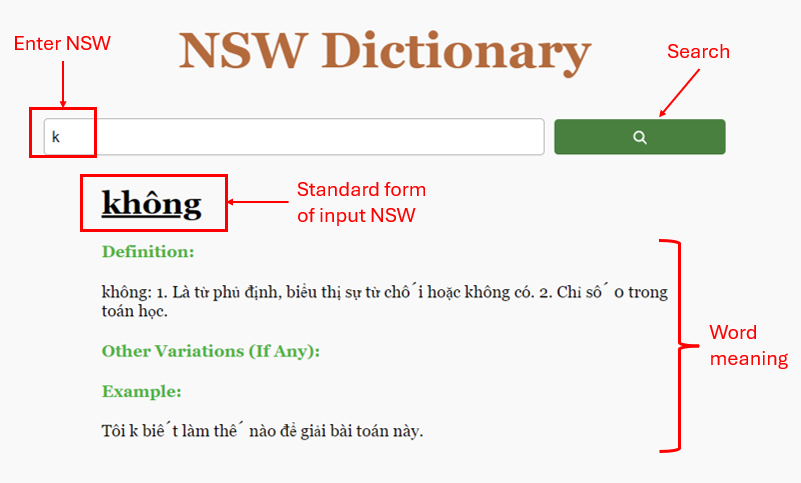}
    \caption{User Interface of NSW LookUp Service. This interface allows users to search for non-standard words and retrieve their standard equivalents, definitions, and examples from the dictionary.}
    \label{fig:ui_dict}
\end{figure}

\begin{figure}
    \centering
    \includegraphics[width=\linewidth]{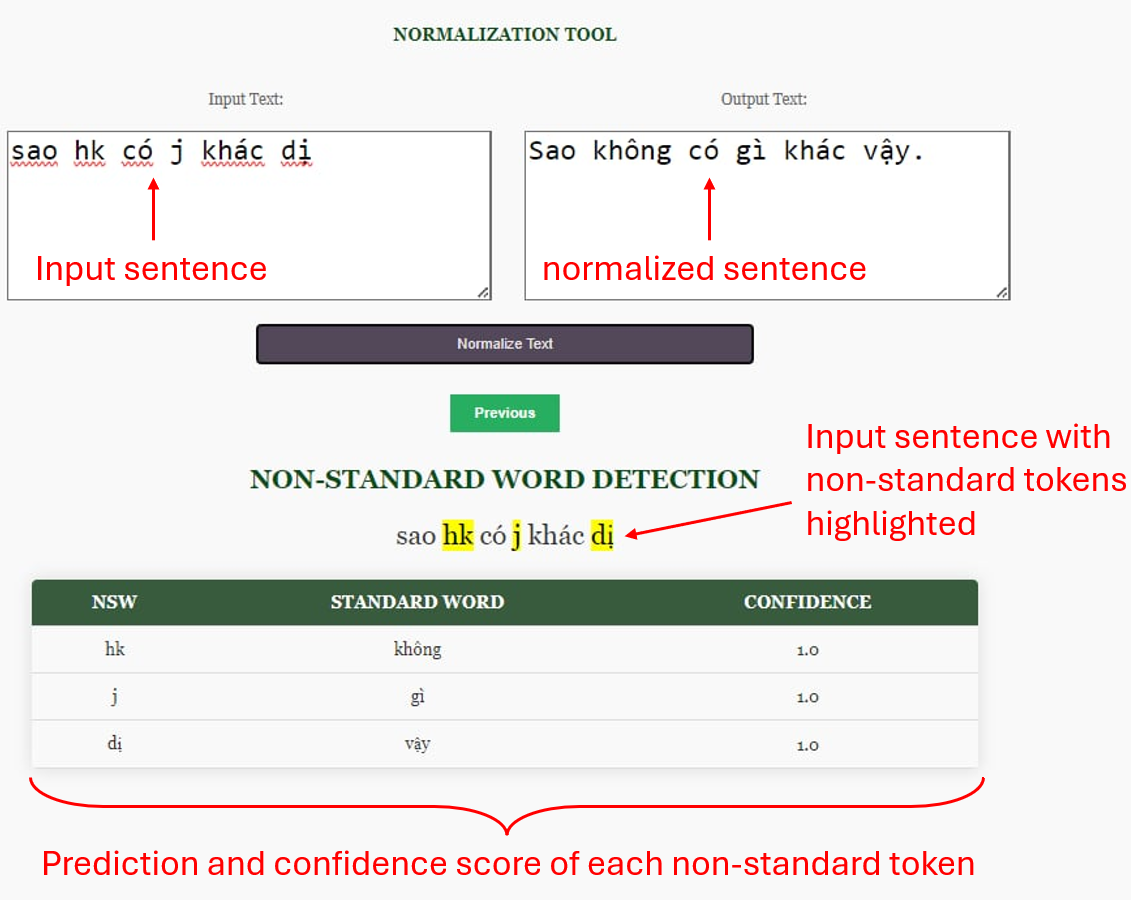}
    \caption{User Interface of Lexical Normalization Service. This interface enables users to input sentences with non-standard words and receive fully normalized outputs in real-time.}
    \label{fig:ui_lexnorm}
\end{figure}

To accommodate non-technical users, we developed a user-friendly front-end interface using a Flask web application. The interface provides two main services, accessible through distinct endpoints:

\begin{itemize}
    \item \textbf{Interactive Dictionary Service}: This service, available via the \texttt{/dict\_lookup} endpoint, allows users to search for non-standard words and retrieve their standard equivalents and definitions from our extensive dictionary. The interface for this service is illustrated in Figure \ref{fig:ui_dict}.
    \item \textbf{Sentence Normalization}: Through the \texttt{/normalize\_text} endpoint, users can input sentences containing non-standard words and receive real-time normalized outputs. The UI for this service is shown in Figure \ref{fig:ui_lexnorm}.
\end{itemize}

The Flask application enables seamless interaction between the front-end and back-end components, ensuring efficient and responsive user experiences. A self-hosting tutorial for deploying the UI is available in the project's GitHub repository, along with a demonstration video on how to use the interface, accessible via this Youtube URL\footnote{\url{https://youtu.be/XBIAogDpF3o?si=PUXiMCuu9qDTfM3B}}.

\section{Discussion and Future Directions}

ViSoLex represents a significant advancement in the lexical normalization of Vietnamese social media text, offering researchers and developers a powerful and flexible tool to address the challenges posed by non-standard language. The system exhibits robust performance in both NSW detection and normalization, leveraging its integration of pre-trained models and weakly supervised learning. Notably, ViSoLex achieves consistent improvements in F1-score and accuracy across multitask settings when compared to the original framework \citet{nguyen2024weaklysuperviseddatalabeling}, which was exclusively designed for lexical normalization without NSW detection. Through its open-source availability, ViSoLex encourages further research and application in Vietnamese NLP.

Looking ahead, future work on ViSoLex will focus on expanding its capabilities to support more complex non-standard patterns and handling additional languages. Efforts will also be made to enhance the model's adaptability, allowing it to better manage evolving trends in social media language. Furthermore, expanding the NSW dictionary and refining the system's ability to predict social-contextual meanings are promising directions. Lastly, improving user experience through more intuitive front-end interfaces and incorporating additional downstream NLP tasks will enhance ViSoLex’s practical applications in real-world scenarios.

\section*{Acknowledgement}
This research is funded by University of Information Technology-Vietnam National University Ho Chi Minh City under grant number D1-2024-75.

\bibliography{custom}

\end{document}